\pdfoutput=1

\documentclass[11pt]{article}

\usepackage[preprint]{acl}

\usepackage{times}
\usepackage{latexsym}

\usepackage[T1]{fontenc}

\usepackage[utf8]{inputenc}

\usepackage{microtype}

\usepackage{inconsolata}
\usepackage{graphicx}
\usepackage{enumitem} 
\usepackage{pifont}

\usepackage{listings}
\usepackage{xcolor}
\usepackage{amsmath} 
\usepackage{todonotes}
\title{LLM-Ref: Enhancing Reference Handling in Technical Writing \\with Large Language Models}

\author{Kazi Ahmed Asif Fuad \\
  Oregon State University \\
  \texttt{fuadk@oregonstate.edu} \\\And
  Lizhong Chen \\
  Oregon State University\\
  \texttt{chenliz@oregonstate.edu} \\}

\begin{document}
\maketitle
\begin{abstract}
Large Language Models (LLMs) excel in data synthesis but can be inaccurate in domain-specific tasks, which retrieval-augmented generation (RAG) systems address by leveraging user-provided data. However, RAGs require optimization in both retrieval and generation stages, which can affect output quality. In this paper, we present LLM-Ref, a writing assistant tool that aids researchers in writing articles from multiple source documents with enhanced reference synthesis and handling capabilities. Unlike traditional RAG systems that use chunking and indexing, our tool retrieves and generates content directly from text paragraphs. This method facilitates direct reference extraction from the generated outputs, a feature unique to our tool. Additionally, our tool employs iterative response generation, effectively managing lengthy contexts within the language model's constraints. Compared to baseline RAG-based systems, our approach achieves a $3.25\times$ to $6.26\times$ increase in Ragas score, a comprehensive metric that provides a holistic view of a RAG system’s ability to produce accurate, relevant, and contextually appropriate responses. This improvement shows our method enhances the accuracy and contextual relevance of writing assistance tools.
\end{abstract}

\section{Introduction}

Scientific research is fundamental in enriching our knowledge base, tackling real-life challenges, and contributing to the betterment of human lives. Writing clear and precise research articles is crucial for disseminating new findings and innovations to a broad audience, avoiding misunderstandings that could impede progress.  Writing research papers clearly is challenging due to the need to balance complex content with readability, adhere to strict formatting, and synthesize coherently. Writing tools aid researchers by providing advanced grammar and style checks, simplifying data organization, and enhancing argument coherence, making them essential for crafting impactful, high-quality scientific papers with real-world applications.

Large Language Models (LLMs) have significantly advanced natural language processing (NLP) by improving language understanding, generation, and interaction. While they excel in many NLP tasks, they require substantial computational resources and may struggle with specialized tasks without domain-specific knowledge. LLMs often produce inaccurate responses or ‘hallucinations' when handling tasks beyond their training data. Developing an effective writing assistant using LLMs requires fine-tuning with domain-specific data from various fields, a process that demands extensive computational resources and a diverse dataset, making it costly to create a versatile and effective tool for diverse writing challenges.

To mitigate the challenges associated with using LLMs for downstream tasks, Retrieval-Augmented Generation (RAG) ~\cite{lewis2021retrievalaugmented} systems have gained popularity for their capability to integrate external user-specific data. By actively sourcing information from knowledge databases during the generation phase, RAG efficiently tackles the challenge of creating content that may be factually inaccurate~\cite{gao2024retrievalaugmented}. When working with user source data, RAG-based systems usually read the documents in text format which they segment into small chunks. However, determining the appropriate size for chunking presents a challenging problem, as it significantly impacts the quality of the final output generated. To manage the model's context limitations, RAG systems often only consider the top-k context segments, potentially overlooking crucial contextual details. Furthermore, due to their data-processing and retrieval approaches, RAG-based systems fall short of providing comprehensive source references needed for composing research articles. 


In this paper, we present LLM-Ref, a writing assistant tool that helps researchers with enhanced reference extraction while writing articles based on multiple source documents. To address the challenges of existing RAG-based tools, our writing assistant tool preserves the hierarchical section-subsection structure of source documents. Rather than dividing texts into chunks and transforming them into embeddings, our approach directly utilizes the paragraphs from research articles to identify information relevant to specific queries. To efficiently retrieve all the relevant information from the source documents, an LLM is utilized due to their superior performance in finding semantic relevance. Efficient utilization of contexts in paragraphs allows LLM-Ref extract references within the contexts. Furthermore, iterative generation of output response allows handling long context and finer responses. Efficient retrieval and preservation of hierarchical source information enable the listing of comprehensive references, ensuring that users have access to detailed citation details. The proposed LLM-Ref can provide both primary references—the source documents—and secondary references, which are listed in the context paragraphs of the source documents. To the best of our knowledge, no other similar work focuses on providing both primary and secondary references. 

Evaluation results show superior performance of our tool over existing RAG-based systems.  The proposed LLM-Ref demonstrates significant performance improvements over other RAG systems, achieving a $5.5\times$ higher Context Relevancy in the multiple source documents scenario compared to Basic RAG and a $4.7 \times$ higher Context Relevancy in the single source document scenario. Additionally, it delivers an impressive increase in the Ragas Score, outperforming the best alternative by $3.25\times$ in the multiple source documents scenario and $2.65\times$ in the single source document scenario. These results highlight that the proposed tool provides more accurate, relevant, and contextually precise outputs, enhancing the overall utility and reliability of the writing assistance it offers.


\section{Background and Related Works}

Large Language Models (LLMs) have propelled the landscape of natural language processing (NLP), leveraging vast amounts of data to understand, generate, and interact with human language in a deeply nuanced and contextually aware manner. Models like ChatGPT~\cite{openaichatgpt4, brown2020language} and LLaMa~\cite{touvron2023llama} have demonstrated exceptional performance across a wide range of NLP benchmarks~\cite{bubeck2023sparks, hendrycks2021measuring, srivastava2023imitation}, solidifying their role as indispensable tools in both everyday applications and cutting-edge research. However, the remarkable performance of LLMs incurs huge computational costs to train the several billions of parameters of the model on enormous amounts of data~\cite{kaddour2023challenges}. Moreover, unless fine-tuned for domain-specific downstream tasks, the performance of LLMs degrades notably~\cite{kandpal2023large, gao2024retrievalaugmented}. Being transformer-based models~\cite{vaswani2023attention}, LLMs have restrictions on how much input context they can utilize for response generation which affects the quality of the output. Conversely, LLMs with long context lengths fail to relate the content in the middle. Compounding the challenges, LLMs exhibit ‘hallucinations' when tasks require up-to-date information that extends beyond their training data~\cite{zhang2023sirens, kandpal2023large, gao2024retrievalaugmented}. These drawbacks often complicate developing custom downstream applications with LLMs.

Retrieval-Augmented Generation (RAG)~\cite{lewis2021retrievalaugmented} systems address the challenge of generating potentially factually inaccurate content by actively sourcing information from external knowledge databases during the generation phase. The basic workflow of Retrieval-Augmented Generation (RAG) involves several key stages: indexing, retrieval, and generation~\cite{lewis2021retrievalaugmented, ma2023query}. Initially, RAG creates an index from external sources, preparing data through text normalization processes like tokenization and stemming, enhancing searchability. This index is crucial for the subsequent retrieval stage, where models like BERT~\cite{devlin2019bert} enhance accuracy by understanding the semantic nuances of queries. During the final generation phase, the system uses the retrieved information and the initial query to produce relevant and reflective text. This process involves synthesizing the content to ensure it not only aligns with the retrieved data and query intent but also introduces potentially new insights, balancing accuracy with creativity.

Building on the foundational workflow of RAG, recent advancements in large language models (LLMs) have introduced more sophisticated techniques for managing extensive data and enhancing the relevance and accuracy of generated content. MemWalker~\cite{chen2023walking} tackles the limitations of context window size by creating a memory tree from segmented text, which improves indexing and data management for long-context querying. 

This method is complemented by other innovative approaches like KnowledGPT~\cite{wang2023knowledgpt} and Rewrite-Retrieve-Read~\cite{ma-etal-2023-query}, which refine query manipulation through programming and rewriting techniques to better capture user intent. Such approaches suffer from the complexity of multi-hop queries where error propagation affects the response significantly.

In parallel, PRCA~\cite{yang-etal-2023-prca} employs domain-specific abstractive summarization to extract crucial, context-rich information, enhancing the quality of query responses. FiD-light~\cite{hofstätter2022fidlight} introduces a listwise autoregressive re-ranking method that links generated text to source passages, organizing the retrieval and generation process to improve coherence and relevance. Similarly, RECOMP~\cite{xu2023recomp} compresses information into concise summaries, focusing on the most pertinent content for generation, thus streamlining the workflow and improving output quality. These diverse approaches collectively advance LLMs' capabilities in handling extensive, complex data, refining the interaction between retrieval and generation for more accurate, contextually relevant outputs. However, none of the approaches address reference handling. 

GPT-based models are highly effective at paraphrasing, and grammar correction, and also excel in crafting informative paragraphs suitable for research papers. The latest ChatGPT, GPT-4 can conduct question-answering tasks using user-provided data, marking a significant advancement in its functionality. Despite supporting multiple user files as inputs, ChatGPT does not return the specific context utilized in the generation process nor does it offer comprehensive references. Tools like, \textit{txyz.ai}\footnotemark also facilitate academic researchers in understanding complex research papers by providing summaries and answering questions about a particular academic research article.  However, this tool is designed to interact with only a single file at a time compared to a list of research articles that a researcher typically works with when writing a paper. Moreover, it does not generate a comprehensive list of references cited within the article's context. Its incapability to handle multiple files restricts its use in research writing assistance. On the other hand, tools like \textit{wisio.app}\footnotemark and \textit{jenni.ai}\footnotemark leverage generative features akin to ChatGPT for article writing. The most comparable to our tool is \textit{ChatDoc}\footnotemark, which facilitates interaction with multiple source documents, providing the source context and references of primary files. However, it falls short in offering a comprehensive list of secondary references found within the context.

\footnotetext[1]{\href{https://txyz.ai/}{https://txyz.ai/}} \footnotetext[2]{\href{https://wisio.app/}{https://wisio.app/}} \footnotetext[3]{\href{https://jenni.ai/}{https://jenni.ai/}} \footnotetext[4]{\href{https://chatdoc.com/}{https://chatdoc.doc/}}


\section{Architecture of Proposed LLM-Ref}

\begin{figure*}[!ht]
    \centering
    \includegraphics[width=\textwidth]{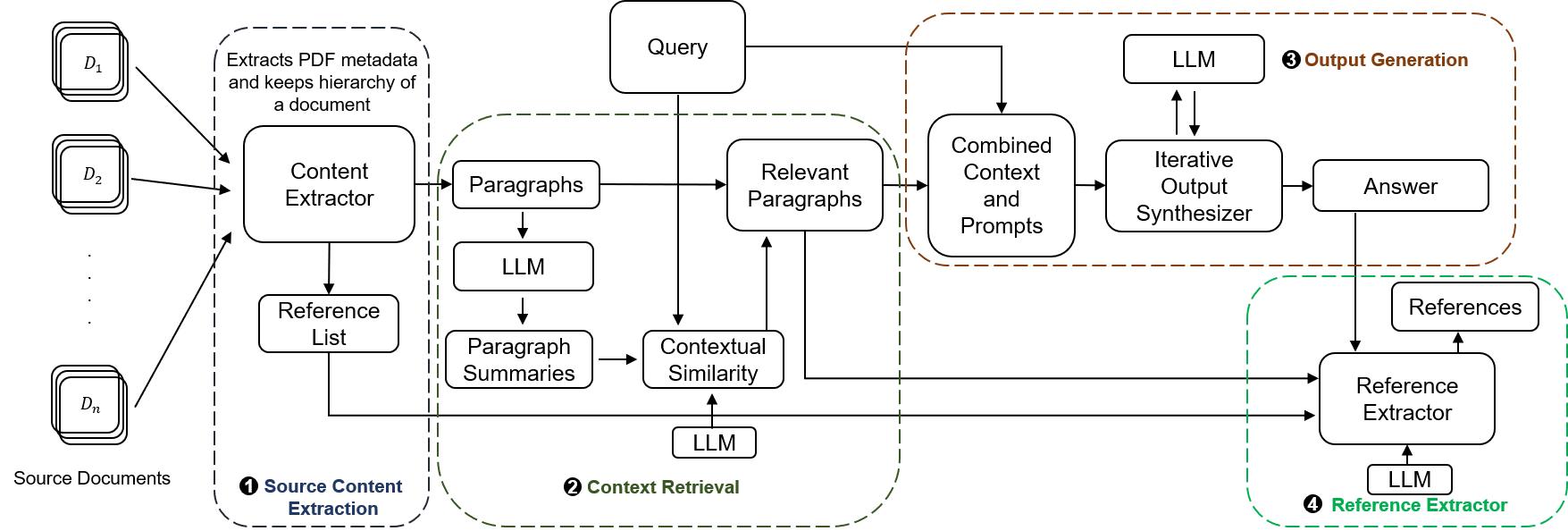}
    \caption{Architecture of the proposed LLM-Ref. \ding{172} \textit{Content Extractor} extracts texts and references, preserving the paragraph hierarchy of each article. Each article metadata along with respective paragraph summaries extracted from LLM is stored offline. For a given \textit{query}, in \ding{173} \textit{Context Retrieval}, relevant paragraphs are extracted and combined with prompts to generate answers. The \ding{174} \textit{Iterative Output Synthesizer} feeds the combined prompt and context to LLM for output text generation based on context length limit. Finally, the \ding{175}  \textit{Reference  Extractor} extracts respective references for output text from relevant paragraphs.}
    \label{fig:block_diagram}
\end{figure*}

In this section, we propose LLM-Ref, a writing tool designed to assist researchers by providing enhanced reference synthesis and handling capabilities, while synthesizing responses based on the information found within the context of provided research articles. Most RAG-based systems face challenges in the retrieval of adequate and correct input contexts and do not provide source or secondary references when synthesizing results from multiple source documents. In contrast, the proposed LLM-Ref extracts a hierarchical flow of contents in the source documents and provides proper references with the synthesized output. The overall architecture of the system is depicted in Figure~\ref{fig:block_diagram}.

A research article is typically structured into sections and subsections to present and elucidate a particular problem, background information, and analysis. Inside a section or subsection, each paragraph conveys a specific context. As to develop a writing assistant for research articles it is crucial to extract source contents efficiently with proper hierarchy.  Given this, the proposed LLM-Ref begins with \ding{172} Content Extractor by extracting text and references from documents, ensuring the original organization into paragraphs is kept intact. It stores information about each document, including summaries of paragraphs generated by an LLM, in an offline repository. For any particular query, \ding{173} Context Retrieval finds and compiles relevant sections of text, augmenting these with guiding questions to assist in synthesizing responses. A specialized component, \ding{174} Iterative Output Synthesizer then processes this compiled information, using a language model to generate text based on the given input and predefined context length. In the final step, accurate citations are extracted from the context for the synthesized output by \ding{175} Reference Extractor. All the prompts utilized in our work are given in the Appendix~\ref{promptdesigns}.

\subsection{Source Content Extraction}

RAG systems often process source documents as plain text, overlooking section and sub-section-level abstraction.  Capturing this abstraction necessitates machine learning-based text classification and segmentation, relying on domain-specific research article datasets. Although identifying sections or sub-sections is challenging, the consistent styles and formats of research articles reveal document hierarchy. Thus, we leverage text formatting to understand a source document's abstraction.

Our text extractor, \textit{Content Extractor}, reads each PDF file and extracts its contents while maintaining the abstraction of the content flow, utilizing the Python library \texttt{pdfminer}. This library offers fine-grained access to most content objects, allowing the \textit{Content Extractor} to understand the research writing template. First, \textit{Content Extractor} extracts the page layout and font-related statistics from all the pages in a document to identify article formatting details, such as the number of columns and font attributes (name, size, and style). Section and subsection labels are identified by searching for common keywords like `Introduction', `Abstract', `References', `2.1', `3.1', `4.1', `a.', `(a)', etc. However, keyword searching alone is not sufficient to accurately position and extract sections or subsections due to multiple possible instances of same section or subsection name. For precise positioning and extraction, we verify the position and text details of each search item against the formatting details initially acquired. Once the sections and subsections labels are accurately extracted, the text organized in paragraphs is extracted. To identify paragraph separation, we leverage indentation, line spacing, and column information. Thus, we store paragraphs within each section and subsection, preserving the correct abstraction.

In general, RAGs process and store documents by dividing them into chunks and applying embeddings. These embeddings are indexed and later used to retrieve relevant chunks through a similarity operation that compares the input chunks with the query. On the contrary, in our approach, we store source information offline in existing paragraphs. To retrieve relevant context, we additionally store concise and informative summaries of each paragraph which are used in the retrieval stage. However, we utilize corresponding original paragraphs for output generation and reference extraction.

\subsection{Context Retrieval}

In conventional RAG systems, optimal text chunking is crucial for converting text chunks into vector embeddings for similarity operations and retrieval, ensuring accuracy and relevance despite language model context limitations. Optimal chunking, which depends on content type, embedding model specs, query complexity, and application use, is important as overly large or small chunks can lead to sub-optimal results.  Fine-tuning embedding models for specific tasks is essential to align with user queries and content relevance, as generic models may not meet domain-specific needs.

To mitigate the existing challenges in the retrieval stage, we perform contextual similarity between the query and the summarized paragraphs of the source documents using an LLM. The prompt consists of the user query and a paragraph from a source document. Once the relevant paragraphs are identified using the corresponding summaries, the original paragraphs are selected and fed as context for the output generation step. In our experiments, LLM-based contextual similarity performs better than embedding-based approaches due to their superior performance in understanding underlying context. Although overlapping or sliding window-based large chunking positively affects retrieving contexts, LLM-based contextual similarity on paragraphs has a better outcome on output generation and reference extraction. Using paragraphs as context can be challenging due to the LLM's context length limitations, a problem we mitigate with our iterative output generation step should it arise.

\subsection{Output Generation}

In the output generation step, the user query and the relevant context paragraphs are combined and fed to the LLM. Usually, it is observed that research paper-related queries tend to have many context paragraphs which often do not fit within the context limit of the LLM. Moreover, LLM suffers from the ‘Lost in the Middle' phenomenon when the context is too long. To address these issues, the \textit{Iterative Output Synthesizer} is capable of synthesizing responses iteratively by processing input paragraphs and ensuring they fit within the context limit of the language model. Initially, the unit feeds the first paragraph (as context) along with the query to an LLM to generate output. The response from the LLM is then continuously updated with the rest of the relevant paragraphs. While the system generates output through continuous updates, it enforces the context limit by monitoring the size of the query, the paragraphs, and the response.

\subsection{Reference Extraction}

Despite their popularity, RAG-based  systems fall short in offering citations. While ChatGPT-4 now has the capability to process user data, it does not provide definite necessary contexts or references that are essential for academic research. In our tool, we extract the references from input context paragraphs.  Our system adeptly identifies the source documents, referred to as ‘primary references', along with the citations found within the source context paragraphs, which we term ‘secondary references'. During the generation phase, LLMs omit citation notations, posing challenges in reference extraction. So our system adopts two presentations of references: Coarse-grain references for broader citation identification and Fine-grain references for more detailed citation tracking. Most research papers use either {\lq{enumerated}\rq} (e.g., {\lq{[1]}\rq}, {\lq{[2-5]}\rq}, {\lq{[3,9]}\rq}) or {\lq{named}\rq} (e.g., {\lq{(Author name et al., 2024)}\rq}) reference notations and our reference extractor is adept at recognizing both types within the contexts. Our reference extraction method can integrate with existing RAGs but requires optimization during the chunking phase.

\subsubsection{Coarse-grain References}

In coarse-grain reference extraction, the \textit{Reference Extractor} catalogs all the references identified within the contexts. As contexts are extracted as paragraphs containing information relevant to the queries, this approach offers a comprehensive overview of a specific issue. The tool enumerates all the source papers and secondary references found within these context paragraphs, thereby furnishing users with extensive details for their assessment and comprehension.

\subsubsection{Fine-grain References}
In fine-grain reference extraction, the \textit{Reference Extractor} meticulously identifies the context lines most relevant to each line in the output text with the help of a LLM. This method of pinpointing the most pertinent context lines enables us to discover more specific references, thus achieving greater precision in our reference extraction process. We determine the highest relevance between response lines and source context lines using an LLM. By identifying the most relevant source contexts, we can extract primary and secondary references with high precision. This process facilitates the rapid compilation of synthesized outputs from a multitude of source documents.

\section{Experimental Setup} 

\subsection{Evaluating RAG Approaches}
Our evaluation compares LLM-Ref with three other RAG implementations: Basic RAG~\cite{lewis2021retrievalaugmented}, Parent-Document Retriever (PDR) RAG~\cite{langchain_pdr_retriever}, and Ensemble RAG~\cite{langchain_ensemble_retriever}, highlighting their methodologies and applications. In all of our experiments, the GPT-3.5 16k model was utilized at all stages of RAG systems.

The Basic RAG approach integrates a retriever and a language model to answer questions based on retrieved documents. It involves splitting documents into chunks, embedding them with models, and storing them in a vector database. The retriever fetches relevant chunks based on the query, which the language model uses to generate accurate responses. 

The PDR RAG enhances retrieval precision by structuring documents into parent-child relationships. Larger parent chunks and smaller child chunks are embedded and stored in a vector database and in-memory store. A ParentDocumentRetriever fetches relevant chunks, providing refined context to the language model, ensuring more precise context and accurate responses.

The Ensemble RAG combines multiple retrievers to leverage their strengths, resulting in a more robust retrieval system. It uses different retrievers, such as BM25 for keyword matching and vector-based retrievers for semantic similarity. An EnsembleRetriever balances their contributions, using the aggregated context for the language model to generate responses, enhancing retrieval robustness and accuracy for complex queries.

\subsection{Dataset}
The evaluation of systems similar to RAG necessitates human-annotated ground truth answers for a variety of questions, a requirement that proves difficult to fulfill across multiple domains. To address this challenge, Ragas~\cite{es2023ragas} and ARES~\cite{saadfalcon2023ares} employ datasets generated by ChatGPT as ground truth from specific documents. We follow this approach by leveraging GPT-4, simulating an advanced researcher, to create research question-answer-context pairs based on the provided source documents. These generated question-answer-context pairs serve as a benchmark to assess the relevance and accuracy of contexts retrieved and outputs generated by RAG, facilitating a comprehensive analysis of evaluation metrics in conjunction with Ragas.

To evaluate our system on domain-specific tasks, we curated a diverse arXiv dataset with question-answer-context pairs from Physics, Mathematics, Computer Science, Quantitative Finance, Electrical Engineering and Systems Science, and Economics.

Our dataset is divided into two subsets for thorough evaluation:

\begin{enumerate}
    \item Multiple Source Document Subset: This subset contains 955 question-answer-context pairs derived from multiple documents within the same subject area. By combining information from various sources, we aim to capture a broader and more comprehensive understanding of each subject.
    \item Single Source Document Subset: This subset includes 544 question-answer-context pairs, each generated from an individual source document. This allows us to assess the system's performance when relying on a single source of information.
\end{enumerate}

During the evaluation, source documents corresponding to the question-answer-context pairs are provided to the RAG systems.


\begin{table*}[!ht]
\centering
\resizebox{\textwidth}{!}{
\begin{tabular}{clllllllll}
\hline
  \multicolumn{1}{c}{\textbf{Name}} &
  \multicolumn{1}{c}{\textbf{\begin{tabular}[c]{@{}c@{}}Answer \\ Relevancy\end{tabular}}} &
  \multicolumn{1}{c}{\textbf{\begin{tabular}[c]{@{}c@{}}Answer \\ Correctness\end{tabular}}} &
  \multicolumn{1}{c}{\textbf{\begin{tabular}[c]{@{}c@{}}Answer \\ Similarity\end{tabular}}} &
  \multicolumn{1}{c}{\textbf{\begin{tabular}[c]{@{}c@{}}Context \\ Relevancy\end{tabular}}} &
  \multicolumn{1}{c}{\textbf{\begin{tabular}[c]{@{}c@{}}Context \\ Precision\end{tabular}}} &
  \multicolumn{1}{c}{\textbf{\begin{tabular}[c]{@{}c@{}}Context \\ Recall\end{tabular}}} &
  \multicolumn{1}{c}{\textbf{\begin{tabular}[c]{@{}c@{}}Faith \\ fulness\end{tabular}}} &
  \multicolumn{1}{c}{\textbf{\begin{tabular}[c]{@{}c@{}}Ragas \\ Score\end{tabular}}} \\ \hline
  \multicolumn{8}{c}{Multiple Source Documents}
        \\ \hline
 Basic RAG    & 0.598 & 0.448 & 0.892 & 0.049 & 0.857 & 0.697 & 0.547 & 0.158   \\
 PDR RAG      & 0.575 & 0.458 & 0.896 & 0.023 & 0.852 & 0.716 & 0.622 & 0.082 \\
 Ens. RAG & 0.613 & 0.459 & 0.905 & 0.043 & 0.851 & 0.717 & 0.600 & 0.143 \\
 LLM-Ref  & 0.948 & 0.568 & 0.942 & 0.268 & 0.976 & 0.705 & 0.629 & 0.513 
        \\ \hline
        \multicolumn{8}{c}{Single Source Document}
        \\ \hline
Basic RAG    & 0.748 & 0.525 & 0.915 & 0.058 & 0.983 & 0.824 & 0.732 & 0.189  \\
 PDR RAG      & 0.729 & 0.523 & 0.915 & 0.024 & 0.980 & 0.858 & 0.748 & 0.089 \\
 Ens. RAG & 0.789 & 0.541 & 0.931 & 0.035 & 0.999 & 0.885 & 0.778 & 0.125 \\
 LLM-Ref  & 0.947 & 0.596 & 0.930 & 0.272 & 0.969 & 0.703 & 0.547 & 0.501
 \\ \hline
\end{tabular}
}
\caption{Metric evaluation result comparison of LLM-Ref with Basic RAG, Parent Document Retriever RAG, and Ensemble Retrieval RAG, using GPT-3.5 as the LLM. A higher metric value indicates better performance.}
\label{tab:my-table}
\end{table*}


\subsection{Evaluation Metrics}
We employ the Ragas~\cite{es2023ragas} framework to evaluate the performance of the RAG systems.
\textit{Faithfulness} ensures the generated response is based on the provided input context, avoiding false or misleading information ({\lq{hallucinations}\rq}). It is crucial for transparency and accuracy, ensuring the context serves as solid evidence for the answer.

\textit{Answer Relevance} measures how well the generated response directly addresses the question, ensuring responses are on-topic and accurately meet the query's requirements. \textit{Answer Similarity} measures how closely the generated answer aligns with the ground truth in both content and intent, reflecting the RAG system's understanding of the concepts and context~\cite{es2023ragas}.

\textit{Context Relevance} ensures the retrieved context is precise and minimizes irrelevant content, which is crucial due to the costs and inefficiencies associated with processing lengthy passages through LLMs, especially when key information is buried in the middle~\cite{liu2023lost}. \textit{Context Precision} gauges the system's ability to prioritize relevant items, ensuring that the most pertinent information is presented first and distinguishing it from irrelevant data. \textit{Context Recall} measures the model's ability to retrieve all relevant information, balancing true positives against false negatives, to ensure no key details are missed.~\cite{es2023ragas}.

The Ragas score combines key metrics: faithfulness, answer relevancy, context relevancy, and context recall~\cite{langchain_ragas_score}. By integrating these metrics, the Ragas score provides a holistic view of a RAG system's ability to produce accurate, relevant, and contextually appropriate responses, guiding improvements for enhanced performance. A comprehensive explanation of the calculations is provided in the Appendix~\ref{ragas_appendix}.


\section{Results and Analysis} 

\begin{figure*}[!ht]
\centering
\begin{tabular}{p{15.6cm}}
\hline
\footnotesize
Query: Pros and Cons of Discrete vs Cascaded Systems \\ \hline

\footnotesize Answer: Discrete units in direct speech-to-speech translation (S2ST) systems offer several advantages over cascaded systems [1-8]. Firstly, they have lower computational costs and inference latency because they require fewer decoding steps [1, 8, 10-12]. This makes them more efficient and faster in translating speech [1-3, 7, 8, 10-12]. Additionally, direct S2ST is a natural approach for languages that do not have a writing system, as it eliminates the need for intermediate text representation. Another benefit of direct S2ST systems with discrete units is their flexibility in training. They can be trained with or without text transcripts, making them suitable for both written and unwritten languages [1, 6, 7, 9, 13]. This versatility allows for a wider range of applications and language support. However, it is important to note that there is a performance gap between direct S2ST systems with discrete units and cascaded systems. Cascaded systems, which combine speech-to-text translation (S2T) and text-to-speech (TTS), may have an edge in terms of translation accuracy [1-3, 6, 7, 9, 13]. The paragraph does not provide specific details about the extent of this performance gap, but it suggests that cascaded systems may outperform direct S2ST systems in terms of translation quality. In summary, discrete units in direct S2ST systems offer advantages such as lower computational costs, support for languages without a writing system, and flexibility in training. However, there is a performance gap compared to cascaded systems, which may have better translation accuracy [1-3].\\ \hline

\footnotesize
References:
\begin{enumerate}[nosep, label={[{\arabic*}]}]
  \item A. Lee et al., “Direct speech-to-speech translation with discrete units,” arXiv, 2021.
  \item A. Lee et al., “Textless speech-to-speech translation on real data,” arXiv, 2021.
  \item S. Popuri et al., “Enhanced Direct Speech-to-Speech Translation Using Self-supervised Pre-training and Data Augmentation,” arXiv, 2022.
  \item Ye Jia et al., Direct speech-to-speech translation with a sequence-to-sequence model. Proc. Interspeech 2019.
  \item Ye Jia et al., Translatotron 2: Robust direct speech-to-speech translation. arXiv 2021.
  \item Lavie et al., “JANUS-III: Speech-to-speech translation in multiple languages.” In 1997 IEEE International Conference on Acoustics, Speech, and Signal Processing.
  \item S. Nakamura, The ATR multilingual speech-to-speech translation system. IEEE Transactions on Audio, Speech, and Language Processing, 2006.
  \item W. Hsu, HuBERT: Self-supervised speech representation learning by masked prediction of hidden units. arXiv preprint arXiv:2106.07447.
  \item C. Zhang, X. Tan et al., “UWSpeech: Speech to speech translation for unwritten languages,” arXiv:2006.07926, 2020.
  \item Q. T. Do, et al., “Toward expressive speech translation: A unified sequence-to-sequence LSTMs approach for translating words and emphasis.” In INTERSPEECH, 2017.
  \item P. D. Aguero, et al., “Prosody generation for speech-to-speech translation.” In 2006 IEEE International Conference on Acoustics Speech and Signal Processing Proceedings, volume 1, pages I–I, 2006.
  \item G. K. Anumanchipalli et al., “Intent transfer in speech-to-speech machine translation.” In 2012 IEEE Spoken Language Technology Workshop (SLT), 2012.
  \item A. Tjandra et al., “Speech-to-speech translation between untranscribed unknown languages.” In 2019 IEEE Automatic Speech Recognition and Understanding Workshop (ASRU), 2019.
\end{enumerate} \\ \hline
\end{tabular}
\caption{Fine-grained reference samples generated by LLM-Ref when GPT-3.5 is used as the LLM.}
\label{tab:fine}
\end{figure*}

\subsection{Metric Analysis} 

Table~\ref{tab:my-table} compares the performance metrics of LLM-Ref with Basic RAG, PDR RAG, and Ens. RAG  in tasks involving both multiple and single source documents, using GPT-3.5 as the LLM. Further analysis with different LLMs is provided in Appendix ~\ref{gpt4omini_res} and ~\ref{ablation_Study}.

In the case of multiple source documents, LLM-Ref significantly outperforms the other methods across most metrics. It achieves an Answer Relevancy score of \(0.948\), substantially higher than Basic RAG (\(0.598\)), PDR RAG (\(0.575\)), and Ens. RAG (\(0.613\)), indicating its effectiveness in providing pertinent and aligned answers to the questions. Its Answer Correctness is \(0.568\), surpassing others ranging from \(0.448\) to \(0.459\), demonstrating superior accuracy. LLM-Ref also attains the highest Answer Similarity score of \(0.942\) compared to others between \(0.892\) and \(0.905\). These metrics based on the final responses demonstrate the superior efficacy of LLM-Ref in generating answers that are well-aligned with the queries and underlying intent. For Context Relevancy and Precision, LLM-Ref scores \(0.268\) and \(0.976\) respectively, are significantly higher than the other methods, which indicates its exceptional ability to retrieve and utilize relevant information. While Context Recall scores are similar across all methods, LLM-Ref achieves the highest Faithfulness score at \(0.629\), showing that its answers are well-grounded in the provided context. The composite Ragas Score for LLM-Ref is \(0.513\), notably higher than Basic RAG (\(0.158\)), PDR RAG (\(0.082\)), and Ens. RAG (\(0.143\)), highlighting its overall effectiveness in multi-document scenarios.

In single-source document tasks, while LLM-Ref maintains strong performance in certain metrics, it exhibits a moderate reduction in others. It maintains the highest Answer Relevancy (\(0.947\)) and Answer Correctness (\(0.596\)), indicating a consistent ability to provide relevant and accurate answers. Its Answer Similarity is \(0.930\), comparable to Ens. RAG (\(0.931\)) and higher than Basic RAG and PDR RAG (both at \(0.915\)). However, LLM-Ref's Context Precision decreases to \(0.969\), slightly lower than others (\(0.980\) to \(0.999\)), and its Context Recall drops to \(0.703\), substantially lower than the others (\(0.824\) to \(0.885\)). This suggests it retrieves less relevant context from a single document. The Faithfulness score also decreases to \(0.547\), lower than Basic RAG (\(0.732\)), PDR RAG (\(0.748\)), and Ens. RAG (\(0.778\)), indicating its answers may be less grounded in the retrieved context. Despite this, LLM-Ref achieves a Ragas Score of \(0.501\), ranging from $2.65 \times$ to $5.63 \times$  higher that of other methods, highlighting its effectiveness in generating relevant, accurate, and consistent answers from both single and multiple-source documents.

The variations in LLM-Ref's performance, particularly in single-source scenarios, may be due to its optimization for multi-document retrieval. When limited to a single document, it may not fully adjust its retrieval strategies, leading to less comprehensive context extraction and lower Faithfulness. While ChatGPT-4 may incorporate its existing knowledge when generating answers in the ground truth dataset, our system relies exclusively on the provided context. In multi-document scenarios, a broader range of contexts enhances faithfulness and context recall.

LLM-Ref consistently retrieves more relevant information and provides precise context compared to other RAG systems, excelling in delivering accurate and consistent answers. This performance, particularly with multiple-source documents, demonstrates its superiority in generating reliable and high-quality responses. Although LLM-Ref maintains high Answer Relevancy and Correctness in single-source contexts, its lower Context Recall and Faithfulness suggest room for improvement in leveraging specific content.

\subsection{Reference Extraction}
To demonstrate the effectiveness of LLM-Ref, we present a sample of the fine-grain references in Figure~\ref{tab:fine}. For the specific query, we generate fine-grained references where LLM-Ref identifies both enumerated and named references such as ‘[11, 12]’ and ‘(Jia et al.,  2021)’ respectively. For better presentation, we list the references in enumerated format here. In the example, we utilized three source documents to generate the response where ‘[1]’, ‘[2]’ and ‘[3]’ are the primary three source references and the rest of ‘[4] - [13]’ are the secondary references found in the primary source references.


\section{Conclusion}
We present a novel writing assistant that can assist researchers in the extraction of relevant references while synthesizing information from source documents. The proposed system can alleviate the challenging optimization required in RAGs and generate output responses effectively. Moreover, our system can list primary and secondary references to assist researchers where in paying more attention to literature investigation. We intend to explore the opportunities of offline open-source LLMs to build a more flexible system in the future.

\section{Limitations and Ethical Considerations}

Our contribution to this work begins with the PDF file reading component, the Content Extractor, which is designed to handle the most common template styles of research articles. The extraction process is based on various heuristics; however, our \textit{Content Extractor} may not efficiently handle all template styles. Extracting references, particularly reference lists, presents challenges that limit the support capabilities of LLM-Ref. We extract reference lists and store them with their identifiers in the texts. Our system has been tested with various research paper templates. It has demonstrated proficiency in successfully extracting context, especially when reference styles are enumerated (e.g., [1], [2], [4, 28]) or named (author et al., year). We developed this writing assistant tool primarily to guide researchers in exploring different aspects of research, rather than to enable the writing of a research article overnight without in-depth investigation. Both our coarse-grain and fine-grain reference extraction methods can guide researchers on where to focus their efforts more intensively.

In this paper, we present the evaluation of our system using GPT models (GPT-3.5 and 4o-mini, as detailed in the appendix). Additionally, we apply our writing assistant tool to the Llama and Claude models, demonstrating similar results, which underscores the efficacy of our approach across a broad range of LLMs. We plan to extend our comprehensive evaluation of the tool across diverse domain-specific research articles, utilizing open-source Large Language Models (LLMs). Given that LLM-Ref leverages the ChatGPT API, mitigating model bias poses a significant challenge. To minimize potential bias in responses, several measures have been implemented. Specifically, when generating responses to a query, only the contexts identified within the relevant uploaded PDF files are used. Furthermore, the ‘temperature' parameter is set to zero, thereby eliminating randomness in the generation process. This approach helps to maintain that the generated responses are closely aligned with the input contexts and maintain a high degree of specificity.

\bibliography{custom}

\begin{thebibliography}{25}
\expandafter\ifx\csname natexlab\endcsname\relax\def\natexlab#1{#1}\fi

\bibitem[{Brown et~al.(2020)Brown, Mann, Ryder, Subbiah, Kaplan, Dhariwal,
  Neelakantan, Shyam, Sastry, Askell, Agarwal, Herbert-Voss, Krueger, Henighan,
  Child, Ramesh, Ziegler, Wu, Winter, Hesse, Chen, Sigler, Litwin, Gray, Chess,
  Clark, Berner, McCandlish, Radford, Sutskever, and
  Amodei}]{brown2020language}
Tom~B. Brown, Benjamin Mann, Nick Ryder, Melanie Subbiah, Jared Kaplan,
  Prafulla Dhariwal, Arvind Neelakantan, Pranav Shyam, Girish Sastry, Amanda
  Askell, Sandhini Agarwal, Ariel Herbert-Voss, Gretchen Krueger, Tom Henighan,
  Rewon Child, Aditya Ramesh, Daniel~M. Ziegler, Jeffrey Wu, Clemens Winter,
  Christopher Hesse, Mark Chen, Eric Sigler, Mateusz Litwin, Scott Gray,
  Benjamin Chess, Jack Clark, Christopher Berner, Sam McCandlish, Alec Radford,
  Ilya Sutskever, and Dario Amodei. 2020.
\newblock \href {http://arxiv.org/abs/2005.14165} {Language models are few-shot
  learners}.

\bibitem[{Bubeck et~al.(2023)Bubeck, Chandrasekaran, Eldan, Gehrke, Horvitz,
  Kamar, Lee, Lee, Li, Lundberg, Nori, Palangi, Ribeiro, and
  Zhang}]{bubeck2023sparks}
Sébastien Bubeck, Varun Chandrasekaran, Ronen Eldan, Johannes Gehrke, Eric
  Horvitz, Ece Kamar, Peter Lee, Yin~Tat Lee, Yuanzhi Li, Scott Lundberg,
  Harsha Nori, Hamid Palangi, Marco~Tulio Ribeiro, and Yi~Zhang. 2023.
\newblock \href {http://arxiv.org/abs/2303.12712} {Sparks of artificial general
  intelligence: Early experiments with gpt-4}.

\bibitem[{Chen et~al.(2023)Chen, Pasunuru, Weston, and
  Celikyilmaz}]{chen2023walking}
Howard Chen, Ramakanth Pasunuru, Jason Weston, and Asli Celikyilmaz. 2023.
\newblock \href {http://arxiv.org/abs/2310.05029} {Walking down the memory
  maze: Beyond context limit through interactive reading}.

\bibitem[{Devlin et~al.(2019)Devlin, Chang, Lee, and
  Toutanova}]{devlin2019bert}
Jacob Devlin, Ming-Wei Chang, Kenton Lee, and Kristina Toutanova. 2019.
\newblock \href {http://arxiv.org/abs/1810.04805} {Bert: Pre-training of deep
  bidirectional transformers for language understanding}.

\bibitem[{Es et~al.(2023)Es, James, Espinosa-Anke, and
  Schockaert}]{es2023ragas}
Shahul Es, Jithin James, Luis Espinosa-Anke, and Steven Schockaert. 2023.
\newblock \href {http://arxiv.org/abs/2309.15217} {Ragas: Automated evaluation
  of retrieval augmented generation}.

\bibitem[{Gao et~al.(2024)Gao, Xiong, Gao, Jia, Pan, Bi, Dai, Sun, Guo, Wang,
  and Wang}]{gao2024retrievalaugmented}
Yunfan Gao, Yun Xiong, Xinyu Gao, Kangxiang Jia, Jinliu Pan, Yuxi Bi, Yi~Dai,
  Jiawei Sun, Qianyu Guo, Meng Wang, and Haofen Wang. 2024.
\newblock \href {http://arxiv.org/abs/2312.10997} {Retrieval-augmented
  generation for large language models: A survey}.

\bibitem[{Hendrycks et~al.(2021)Hendrycks, Burns, Basart, Zou, Mazeika, Song,
  and Steinhardt}]{hendrycks2021measuring}
Dan Hendrycks, Collin Burns, Steven Basart, Andy Zou, Mantas Mazeika, Dawn
  Song, and Jacob Steinhardt. 2021.
\newblock \href {http://arxiv.org/abs/2009.03300} {Measuring massive multitask
  language understanding}.

\bibitem[{Kaddour et~al.(2023)Kaddour, Harris, Mozes, Bradley, Raileanu, and
  McHardy}]{kaddour2023challenges}
Jean Kaddour, Joshua Harris, Maximilian Mozes, Herbie Bradley, Roberta
  Raileanu, and Robert McHardy. 2023.
\newblock \href {http://arxiv.org/abs/2307.10169} {Challenges and applications
  of large language models}.

\bibitem[{Kandpal et~al.(2023)Kandpal, Deng, Roberts, Wallace, and
  Raffel}]{kandpal2023large}
Nikhil Kandpal, Haikang Deng, Adam Roberts, Eric Wallace, and Colin Raffel.
  2023.
\newblock \href {http://arxiv.org/abs/2211.08411} {Large language models
  struggle to learn long-tail knowledge}.

\bibitem[{LangChain(2023{\natexlab{a}})}]{langchain_ensemble_retriever}
LangChain. 2023{\natexlab{a}}.
\newblock Ensemble retriever.
\newblock
  \url{https://python.langchain.com/v0.1/docs/modules/data_connection/retrievers/ensemble/}.
\newblock Accessed: 2024-03-13.

\bibitem[{LangChain(2023{\natexlab{b}})}]{langchain_ragas_score}
LangChain. 2023{\natexlab{b}}.
\newblock Evaluating rag pipelines with ragas + langsmith.
\newblock
  \url{https://blog.langchain.dev/evaluating-rag-pipelines-with-ragas-langsmith/}.
\newblock Accessed: 2024-01-12.

\bibitem[{LangChain(2023{\natexlab{c}})}]{langchain_pdr_retriever}
LangChain. 2023{\natexlab{c}}.
\newblock Parent document retriever.
\newblock
  \url{https://python.langchain.com/v0.1/docs/modules/data_connection/retrievers/parent_document_retriever/}.
\newblock Accessed: 2024-03-13.

\bibitem[{Lewis et~al.(2021)Lewis, Perez, Piktus, Petroni, Karpukhin, Goyal,
  Küttler, Lewis, tau Yih, Rocktäschel, Riedel, and
  Kiela}]{lewis2021retrievalaugmented}
Patrick Lewis, Ethan Perez, Aleksandra Piktus, Fabio Petroni, Vladimir
  Karpukhin, Naman Goyal, Heinrich Küttler, Mike Lewis, Wen tau Yih, Tim
  Rocktäschel, Sebastian Riedel, and Douwe Kiela. 2021.
\newblock \href {http://arxiv.org/abs/2005.11401} {Retrieval-augmented
  generation for knowledge-intensive nlp tasks}.

\bibitem[{Liu et~al.(2023)Liu, Lin, Hewitt, Paranjape, Bevilacqua, Petroni, and
  Liang}]{liu2023lost}
Nelson~F. Liu, Kevin Lin, John Hewitt, Ashwin Paranjape, Michele Bevilacqua,
  Fabio Petroni, and Percy Liang. 2023.
\newblock \href {http://arxiv.org/abs/2307.03172} {Lost in the middle: How
  language models use long contexts}.

\bibitem[{Ma et~al.(2023{\natexlab{a}})Ma, Gong, He, Zhao, and
  Duan}]{ma2023query}
Xinbei Ma, Yeyun Gong, Pengcheng He, Hai Zhao, and Nan Duan.
  2023{\natexlab{a}}.
\newblock \href {http://arxiv.org/abs/2305.14283} {Query rewriting for
  retrieval-augmented large language models}.

\bibitem[{Ma et~al.(2023{\natexlab{b}})Ma, Gong, He, Zhao, and
  Duan}]{ma-etal-2023-query}
Xinbei Ma, Yeyun Gong, Pengcheng He, Hai Zhao, and Nan Duan.
  2023{\natexlab{b}}.
\newblock \href {https://doi.org/10.18653/v1/2023.emnlp-main.322} {Query
  rewriting in retrieval-augmented large language models}.
\newblock In \emph{Proceedings of the 2023 Conference on Empirical Methods in
  Natural Language Processing}, pages 5303--5315, Singapore. Association for
  Computational Linguistics.

\bibitem[{OpenAI(2023)}]{openaichatgpt4}
OpenAI. 2023.
\newblock \href {http://arxiv.org/abs/2303.08774} {{GPT-4} {T}echnical
  {R}eport}.

\bibitem[{Saad-Falcon et~al.(2023)Saad-Falcon, Khattab, Potts, and
  Zaharia}]{saadfalcon2023ares}
Jon Saad-Falcon, Omar Khattab, Christopher Potts, and Matei Zaharia. 2023.
\newblock \href {http://arxiv.org/abs/2311.09476} {Ares: An automated
  evaluation framework for retrieval-augmented generation systems}.

\bibitem[{Srivastava et~al.(2023)Srivastava, Rastogi, Rao, Shoeb, Abid, Fisch,
  Brown, Santoro, Gupta, Garriga-Alonso, Kluska, Lewkowycz, Agarwal, Power,
  Ray, Warstadt, Kocurek, Safaya, Tazarv, Xiang, Parrish, Nie, Hussain, Askell,
  Dsouza, Slone, Rahane, Iyer, Andreassen, and et.
  al.}]{srivastava2023imitation}
Aarohi Srivastava, Abhinav Rastogi, Abhishek Rao, Abu Awal~Md Shoeb, Abubakar
  Abid, Adam Fisch, Adam~R. Brown, Adam Santoro, Aditya Gupta, Adrià
  Garriga-Alonso, Agnieszka Kluska, Aitor Lewkowycz, Akshat Agarwal, Alethea
  Power, Alex Ray, Alex Warstadt, Alexander~W. Kocurek, Ali Safaya, Ali Tazarv,
  Alice Xiang, Alicia Parrish, Allen Nie, Aman Hussain, Amanda Askell, Amanda
  Dsouza, Ambrose Slone, Ameet Rahane, Anantharaman~S. Iyer, Anders Andreassen,
  and et. al. 2023.
\newblock \href {http://arxiv.org/abs/2206.04615} {Beyond the imitation game:
  Quantifying and extrapolating the capabilities of language models}.

\bibitem[{Touvron et~al.(2023)Touvron, Lavril, Izacard, Martinet, Lachaux,
  Lacroix, Rozière, Goyal, Hambro, Azhar, Rodriguez, Joulin, Grave, and
  Lample}]{touvron2023llama}
Hugo Touvron, Thibaut Lavril, Gautier Izacard, Xavier Martinet, Marie-Anne
  Lachaux, Timothée Lacroix, Baptiste Rozière, Naman Goyal, Eric Hambro,
  Faisal Azhar, Aurelien Rodriguez, Armand Joulin, Edouard Grave, and Guillaume
  Lample. 2023.
\newblock \href {http://arxiv.org/abs/2302.13971} {Llama: Open and efficient
  foundation language models}.

\bibitem[{Vaswani et~al.(2023)Vaswani, Shazeer, Parmar, Uszkoreit, Jones,
  Gomez, Kaiser, and Polosukhin}]{vaswani2023attention}
Ashish Vaswani, Noam Shazeer, Niki Parmar, Jakob Uszkoreit, Llion Jones,
  Aidan~N. Gomez, Lukasz Kaiser, and Illia Polosukhin. 2023.
\newblock \href {http://arxiv.org/abs/1706.03762} {Attention is all you need}.

\bibitem[{Wang et~al.(2023)Wang, Yang, Qiu, Liang, He, Gu, Xiao, and
  Wang}]{wang2023knowledgpt}
Xintao Wang, Qianwen Yang, Yongting Qiu, Jiaqing Liang, Qianyu He, Zhouhong Gu,
  Yanghua Xiao, and Wei Wang. 2023.
\newblock \href {http://arxiv.org/abs/2308.11761} {Knowledgpt: Enhancing large
  language models with retrieval and storage access on knowledge bases}.

\bibitem[{Xu et~al.(2023)Xu, Shi, and Choi}]{xu2023recomp}
Fangyuan Xu, Weijia Shi, and Eunsol Choi. 2023.
\newblock \href {http://arxiv.org/abs/2310.04408} {Recomp: Improving
  retrieval-augmented lms with compression and selective augmentation}.

\bibitem[{Yang et~al.(2023)Yang, Li, Zhang, Wang, Cheng, Li, and
  Xiao}]{yang-etal-2023-prca}
Haoyan Yang, Zhitao Li, Yong Zhang, Jianzong Wang, Ning Cheng, Ming Li, and
  Jing Xiao. 2023.
\newblock \href {https://doi.org/10.18653/v1/2023.emnlp-main.326} {{PRCA}:
  Fitting black-box large language models for retrieval question answering via
  pluggable reward-driven contextual adapter}.
\newblock In \emph{Proceedings of the 2023 Conference on Empirical Methods in
  Natural Language Processing}, pages 5364--5375, Singapore. Association for
  Computational Linguistics.

\bibitem[{Zhang et~al.(2023)Zhang, Li, Cui, Cai, Liu, Fu, Huang, Zhao, Zhang,
  Chen, Wang, Luu, Bi, Shi, and Shi}]{zhang2023sirens}
Yue Zhang, Yafu Li, Leyang Cui, Deng Cai, Lemao Liu, Tingchen Fu, Xinting
  Huang, Enbo Zhao, Yu~Zhang, Yulong Chen, Longyue Wang, Anh~Tuan Luu, Wei Bi,
  Freda Shi, and Shuming Shi. 2023.
\newblock \href {http://arxiv.org/abs/2309.01219} {Siren's song in the ai
  ocean: A survey on hallucination in large language models}.

\end{thebibliography}

\appendix

\section{Appendix}
\label{sec:appendix}


\subsection{Basic Retrieval-Augmented Generation (RAG)}
Retrieval-Augmented Generation (RAG) is an advanced technique that combines information retrieval with text generation, making it particularly effective when generating responses that require specific contextual information from an external knowledge base. The process is typically divided into three main stages: Ingestion, retrieval, and response generation.

\textbf{\textit{Ingestion}}: Once an input file is read, the first stage in RAG involves chunking and embedding, where source texts are segmented into smaller, manageable units, which are then converted into embedding vectors for retrieval. Smaller chunks generally enhance query precision and relevance, while larger chunks may introduce noise, reducing accuracy. Effective chunk size management is crucial for balancing comprehensiveness and precision. Embedding transforms both the user’s query and knowledge base documents into comparable formats, enabling the retrieval of the most relevant information.

\textbf{\textit{Retrieval}}: In the next stage, the relevant information is retrieved from a vector knowledge base such as FAISS. The retriever searches this vector store to find the most relevant chunks of information based on the user's query. This stage is crucial for ensuring that the model has access to the necessary context for generating accurate and contextually relevant responses.

\textbf{\textit{Response Generation}}: In the final stage, the retrieved context is combined with the user's query and fed into the LLM, such as GPT-4, to generate a coherent and relevant response. The model uses the context provided by the retrieved documents to produce answers that are informed by the most pertinent information available. This step highlights the synergy between retrieval and generation, ensuring that the output is not only accurate but also contextually grounded.

Each stage of the RAG process is designed to leverage the strengths of both retrieval and generation, enabling the creation of responses that are informed by specific and relevant external knowledge. By combining these components, RAG systems can significantly enhance the quality and relevance of generated content, making them a powerful tool for applications requiring precise and contextually aware responses.

\subsection{Our System: LLM-Ref}
\label{appendix_llmref}
In contrast to traditional RAG-based systems, our approach emphasizes preserving the hierarchical structure of source data in research writing, enabling the sequential retrieval of relevant contexts and references. During the ingestion stage, our method eliminates the need for a vector store, allowing extracted source information to be stored either online or offline, thereby enhancing flexibility. In the retrieval stage, we leverage large language models (LLMs) to identify the most relevant context paragraphs corresponding to the user query. This approach is particularly well-suited for research article writing, where our findings indicate that each paragraph typically presents a coherent argument, sufficient for establishing contextual similarity. Embedding-based approaches like FAISS rely on pre-computed vector similarities for similarity search and retrieval, which can lead to a loss of subtle contextual nuances present in the data. In contrast, large language models (LLMs) dynamically process and interpret text to capture complex, nuanced relationships within the text. Finally, in the generation stage, our system iteratively produces and refines the response, ensuring accuracy and relevance. While our approach invokes the LLM multiple times across various stages, the associated financial costs are minimal in the context of overall research expenditures.

Extracting both primary and secondary references from source documents requires the LLM to be deterministic. In research articles, the ability to extract contexts from exact paragraphs is crucial. Our experiments with ChatGPT models, including GPT-3.5 and GPT-4, indicate that while these models can refer to uploaded source documents, their generative nature prevents them from providing exact reproductions of contexts or references from the original sources. As a result, it is challenging to precisely identify specific references or corresponding contexts in the original documents based on ChatGPT's responses.

\subsection{Result and Analysis of GPT-4o mini}
\label{gpt4omini_res}

\subsubsection{Metric Analysis}
Table~\ref{tab:my-table_o1mini} presents a comparison of performance metrics for LLM-Ref, Basic RAG, PDR RAG, and Ens. RAG using GPT-4o-mini as the LLM, in tasks involving both multiple and single-source documents. 

\begin{table*}[!ht] 
	\centering
	\resizebox{\textwidth}{!}{
		\begin{tabular}{clllllllll}
			\hline
			\multicolumn{1}{c}{\textbf{Name}} &
			\multicolumn{1}{c}{\textbf{\begin{tabular}[c]{@{}c@{}}Answer \\ Relevancy\end{tabular}}} &
			\multicolumn{1}{c}{\textbf{\begin{tabular}[c]{@{}c@{}}Answer \\ Correctness\end{tabular}}} &
			\multicolumn{1}{c}{\textbf{\begin{tabular}[c]{@{}c@{}}Answer \\ Similarity\end{tabular}}} &
			\multicolumn{1}{c}{\textbf{\begin{tabular}[c]{@{}c@{}}Context \\ Relevancy\end{tabular}}} &
			\multicolumn{1}{c}{\textbf{\begin{tabular}[c]{@{}c@{}}Context \\ Precision\end{tabular}}} &
			\multicolumn{1}{c}{\textbf{\begin{tabular}[c]{@{}c@{}}Context \\ Recall\end{tabular}}} &
			\multicolumn{1}{c}{\textbf{\begin{tabular}[c]{@{}c@{}}Faith \\ fulness\end{tabular}}} &
			\multicolumn{1}{c}{\textbf{\begin{tabular}[c]{@{}c@{}}Ragas \\ Score\end{tabular}}} \\ \hline
			\multicolumn{8}{c}{Multiple Source Documents}
			\\ \hline
			Basic RAG    & 0.675 & 0.517 & 0.890 & 0.049 & 0.846 & 0.698 & 0.582 & 0.159 \\
			PDR RAG     & 0.557 & 0.465 & 0.861 & 0.034 & 0.828 & 0.587 & 0.590 & 0.116 \\
			Ens. RAG 	& 0.709 & 0.531 & 0.899 & 0.037 & 0.851 & 0.726 & 0.615 & 0.129  \\
			LLM-Ref  & 0.966 & 0.546 & 0.947 & 0.246 & 0.980 & 0.732 & 0.569 & 0.486
			\\ \hline
			\multicolumn{8}{c}{Single Source Document}
			\\ \hline
			Basic RAG    & 0.742 & 0.556 & 0.899 & 0.048 & 0.991 & 0.768 & 0.625 & 0.158  \\
			PDR RAG      & 0.788 & 0.583 & 0.911 & 0.025 & 0.980 & 0.864 & 0.775 & 0.090\\
			Ens. RAG & 0.816 & 0.591 & 0.920 & 0.046 & 0.998 & 0.843 & 0.738 & 0.157\\
			LLM-Ref  & 0.952 & 0.636 & 0.932 & 0.267 & 0.961 & 0.734 & 0.530 & 0.497
			\\ \hline
		\end{tabular}
	}
	\caption{Metric Evaluation result comparison of LLM-Ref with Basic RAG, Parent Document Retriever RAG, and Ensemble Retrieval RAG, using GPT 4o-mini as the LLM. A higher value of a metric indicates better performance.}
	\label{tab:my-table_o1mini}
\end{table*}


In tasks involving multiple source documents, LLM-Ref consistently outperforms the other methods across several key metrics. It achieves the highest Answer Relevancy score of \(0.966\), significantly higher than Basic RAG (\(0.675\)), PDR RAG (\(0.557\)), and Ens. RAG (\(0.709\)), indicating its superior capability to provide relevant answers. Additionally, LLM-Ref's Answer Correctness is \(0.546\), demonstrating improved accuracy over Basic RAG (\(0.517\)) and PDR RAG (\(0.465\)). With the highest Answer Similarity of \(0.947\), LLM-Ref also demonstrates its ability to generate answers closely aligned with the ground truth, outperforming others in the range of \(0.861\) to \(0.899\). In terms of Context Relevancy, LLM-Ref shows significant improvement with a score of \(0.246\), outperforming all other methods, highlighting its ability to retrieve pertinent information. Although Context Recall is slightly lower than Ens. RAG and Basic RAG, the high Context Precision of \(0.980\) and Faithfulness score of \(0.569\) emphasize LLM-Ref's overall reliability in multi-document tasks. Its Ragas score of \(0.486\) further reinforces its robust performance, well beyond Basic RAG (\(0.159\)), PDR RAG (\(0.116\)), and Ens. RAG (\(0.129\)).

In single-source document tasks, LLM-Ref maintains strong results, particularly in Answer Relevancy, where it scores \(0.952\), outpacing other methods such as Basic RAG (\(0.742\)) and Ens. RAG (\(0.816\)). Its Answer Correctness also stands out at \(0.636\), higher than Ens. RAG (\(0.591\)) and Basic RAG (\(0.556\)). LLM-Ref's Answer Similarity remains competitive at \(0.932\), slightly lower than Ens. RAG (\(0.920\)), but still higher than others. While its Context Precision is lower than Ens. RAG (\(0.961\) vs. \(0.998\)), it continues to demonstrate a strong Context Relevancy score of \(0.267\), significantly surpassing other methods. However, LLM-Ref's Context Recall decreases to \(0.734\), lower than Basic RAG (\(0.768\)) and Ens. RAG (\(0.843\)), which suggests that the model retrieves less relevant context in single-document settings. The Faithfulness score for LLM-Ref is \(0.530\), lower than Basic RAG (\(0.625\)) and Ens. RAG (\(0.738\)), indicating room for improvement in grounding answers in the provided context. Nonetheless, LLM-Ref achieves a strong Ragas score of \(0.497\), significantly outperforming Basic RAG (\(0.158\)) and Ens. RAG (\(0.157\)), showcasing its consistent ability to generate accurate and relevant answers in both single and multi-document tasks.

The performance variation in single-source tasks may be attributed to LLM-Ref’s optimization for multi-document retrieval, where it excels in aggregating and leveraging context across multiple sources. In single-source scenarios, it appears that the model may not fully optimize its context retrieval strategies, leading to slightly lower metrics for Context Recall and Faithfulness.

\subsubsection{Computation Costs}
The proposed method is meticulously designed to support the writing of research articles, a task that requires a high degree of precision. Compared to traditional Retrieval-Augmented Generation (RAG) systems, our approach incurs higher computational costs due to its focus on achieving enhanced accuracy. However, leveraging open-source large language models (LLMs) fine-tuned for specific tasks can help mitigate these expenses.

The computational overhead of our system, in contrast to traditional RAG systems, can be articulated as follows:

\begin{enumerate}
    \item \textbf{Content Extraction:} The system generates summaries for each paragraph extracted from the documents, storing these summaries for subsequent context extraction. The number of LLM calls made during this step is equal to the number of paragraphs, denoted as \( N \). Traditional RAG systems typically do not invoke LLMs at this stage, instead generating embeddings and storing them in a vector index.

    \item \textbf{Context Extraction:} During this phase, the LLM is invoked \( N \) times to find relevant paragraphs to the query, utilizing the paragraph summaries to minimize the token count, thereby reducing the computational load.

    \item \textbf{Generation:} The generation of responses is conducted iteratively based on the retrieved contexts. The number of LLM calls in this phase depends on the number of contexts retrieved, denoted as \( c \). Our experiments indicate that LLM-Ref retrieves approximately half the number of contexts compared to traditional RAG systems when all the relevant contexts are chosen, leading to reduced computational demands.

    \item \textbf{Reference Extraction:} This step is unique to our system and involves additional LLM calls, denoted as \( p \times q \), where \( p \) represents the number of lines in the generated response and \( q \) corresponds to the lines present in the context. This process ensures the precision and relevance of the extracted references.
\end{enumerate}

LLM calls in content extraction are executed only once during the initial reading of the document and storage of summaries. However, each query necessitates LLM calls in context extraction, answer generation, and reference extraction.

\noindent Therefore, each query requires \( (N + c + p \times q) \) LLM calls. Assuming we have \( N = 50 \) paragraphs, \( c = 8 \) contexts, \( p = 7 \) generated lines, and \( q = 8 \) lines per context, the total is 56 lines. Additionally, each paragraph contains 220 tokens on average, each line approximately 25 tokens, and prompts contain 60 tokens.
\begin{center}
\begin{align*}
N &= 50 \times (220 + 60) \\ &= 14,000 \, \text{tokens}\\
c &= 8 \times (7 \times 25 + 60) + 1000 \\ &= 2,880 \, \text{tokens} \\
p \times q &= 7 \times 8 \times 7 \\ &= 392 \, \text{LLM calls} \\
\text{Total tokens} &= 14,000 + 2,880 \\
&\quad + 392 \times (25 + 25 + 15) \\ &= 42,360 \, \text{tokens}
\end{align*}
\end{center}
Thus, the total input tokens amount to 42,360 tokens.

During both content extraction and reference extraction, the LLM returns only ‘True' or ‘False' for comparison, producing just one token. However, during generation, as it iteratively generates and refines the response, we estimate approximately 1,500 tokens are generated.

\noindent Output tokens = \( 50 + 1500 + 392 = 1942 \) tokens.

\noindent If we use GPT-4o-mini, which costs \$0.150 per 1M input tokens and \$0.600 per 1M output tokens as of October 2024, the cost per query (CpQ) in USD is calculated as:
\[
\text{CpQ} = \frac{0.150}{10^6} \times 42360 + \frac{0.600}{10^6} \times 1942 \approx 0.0075
\]
\noindent Considering the funds typically allocated to research, the cost of using our proposed LLM-Ref for article writing is minimal. Table~\ref{tab:expenses} provides a detailed account of the actual expenses associated with conducting the experiments outlined in  Table~\ref{tab:my-table_o1mini}.

\noindent In conclusion, while our system incurs higher computational costs, such costs are common in similar applications. Evaluation frameworks like Ragas and ARES, which rely on LLMs to assess similarities, incur similar expenses. In return, LLM-Ref offers enhanced accuracy and precision in content generation, crucial for research article writing.


\begin{table}[!ht]
	\centering
	\resizebox{8cm}{!}{
		\begin{tabular}{cllll}
			\hline
			\multicolumn{1}{c}{\textbf{}} & \multicolumn{1}{c}{\textbf{\begin{tabular}[c]{@{}c@{}}Expense \\ (\$)\end{tabular}}} & \multicolumn{1}{c}{\textbf{\begin{tabular}[c]{@{}c@{}}Input \\ Tokens\end{tabular}}} & \multicolumn{1}{c}{\textbf{\begin{tabular}[c]{@{}c@{}}Output \\ Tokens\end{tabular}}} \\ \hline
			\multicolumn{4}{c}{\textbf{Multiple Source Documents}} \\ \hline
			LLM-Ref    & 10.07   & 62,798,321   & 1,178,650  \\
			Basic RAG  & 0.3     & 2,050,930    & 80,451     \\
			PDR RAG    & 0.24    & 1,973,782    & 66,592     \\
			Ens. RAG   & 0.47    & 3,014,645    & 89,835     \\ \hline
			\multicolumn{4}{c}{\textbf{Single Source Document}} \\ \hline
			LLM-Ref    & 1.9     & 12,166,856   & 156,614    \\
			Basic RAG  & 0.18    & 1,254,023    & 35,453     \\
			PDR RAG    & 0.34    & 2,526,472    & 48,723     \\
			Ens. RAG   & 0.27    & 1,906,283    & 42,500     \\ \hline
		\end{tabular}
	}
	\caption{Comparison of Expense, Input Tokens, and Output Tokens for Multiple and Single Source Documents when GPT-4o-mini is used as the LLM.}
	\label{tab:expenses}
\end{table}



\subsection{Ablation Study}
\label{ablation_Study}
\subsubsection{Performance Analysis on Different LLMs}
Table~\ref{tab:my-table_nlp_all} compares the performance metrics of LLM-Ref against Basic RAG, PDR RAG, and Ens. RAG across various language models, including GPT-3.5, GPT-4o-mini, Llama 3.1-405b, and Claude 3.5 Sonnet. In this experiment, we focus exclusively on the computer science subset of the multi-document dataset. As before, a higher value across the metrics signifies superior performance. The results demonstrate LLM-Ref’s consistent advantage over other methods, particularly in providing more relevant, correct, and similar answers.

\begin{table*}[!ht] 
	\centering
	\resizebox{\textwidth}{!}{
		\begin{tabular}{clllllllll}
			\hline
			\multicolumn{1}{c}{\textbf{Name}} &
			\multicolumn{1}{c}{\textbf{\begin{tabular}[c]{@{}c@{}}Answer \\ Relevancy\end{tabular}}} &
			\multicolumn{1}{c}{\textbf{\begin{tabular}[c]{@{}c@{}}Answer \\ Correctness\end{tabular}}} &
			\multicolumn{1}{c}{\textbf{\begin{tabular}[c]{@{}c@{}}Answer \\ Similarity\end{tabular}}} &
			\multicolumn{1}{c}{\textbf{\begin{tabular}[c]{@{}c@{}}Context \\ Relevancy\end{tabular}}} &
			\multicolumn{1}{c}{\textbf{\begin{tabular}[c]{@{}c@{}}Context \\ Precision\end{tabular}}} &
			\multicolumn{1}{c}{\textbf{\begin{tabular}[c]{@{}c@{}}Context \\ Recall\end{tabular}}} &
			\multicolumn{1}{c}{\textbf{\begin{tabular}[c]{@{}c@{}}Faith \\ fulness\end{tabular}}} &
			\multicolumn{1}{c}{\textbf{\begin{tabular}[c]{@{}c@{}}Ragas \\ Score\end{tabular}}} \\ \hline
			\multicolumn{9}{c}{GPT 3.5}
			\\ \hline
			
            Basic RAG & 0.545 & 0.412 & 0.899 & 0.044 & 0.999 & 0.665 & 0.588 & 0.143 \\ 
            PDR RAG & 0.619 & 0.460 & 0.926 & 0.014 & 0.999 & 0.783 & 0.607 & 0.052 \\ 
            Ens. RAG & 0.629 & 0.471 & 0.936 & 0.027 & 0.999 & 0.775 & 0.624 & 0.097 \\ 
            LLM-Ref & 0.960 & 0.555 & 0.950 & 0.157 & 0.993 & 0.676 & 0.721 & 0.389 \\ 
            \hline
            \multicolumn{9}{c}{GPT 4o-mini} \\ 
            \hline
            Basic RAG & 0.765 & 0.540 & 0.916 & 0.041 & 0.999 & 0.689 & 0.564 & 0.138 \\ 
            PDR RAG & 0.606 & 0.482 & 0.875 & 0.033 & 0.993 & 0.569 & 0.524 & 0.112 \\ 
            Ens. RAG & 0.857 & 0.572 & 0.939 & 0.027 & 0.993 & 0.757 & 0.668 & 0.096 \\ 
            LLM-Ref & 0.953 & 0.575 & 0.951 & 0.179 & 0.999 & 0.683 & 0.640 & 0.413 \\ 
            \hline
            \multicolumn{9}{c}{Llama 3.1-405b} \\ 
            \hline
            Basic RAG & 0.571 & 0.443 & 0.875 & 0.035 & 0.987 & 0.538 & 0.390 & 0.114 \\ 
            PDR RAG & 0.642 & 0.439 & 0.887 & 0.022 & 0.999 & 0.682 & 0.570 & 0.079 \\ 
            Ens. RAG & 0.744 & 0.491 & 0.915 & 0.030 & 0.999 & 0.725 & 0.641 & 0.105 \\ 
            LLM-Ref & 0.958 & 0.556 & 0.950 & 0.112 & 0.987 & 0.650 & 0.564 & 0.300 \\ 
            \hline
            \multicolumn{9}{c}{Claude 3.5 Sonnet} \\ 
            \hline
            Basic RAG & 0.634 & 0.544 & 0.941 & 0.042 & 0.999 & 0.694 & 0.691 & 0.142 \\ 
            PDR RAG & 0.702 & 0.550 & 0.942 & 0.015 & 0.999 & 0.762 & 0.723 & 0.055 \\ 
            Ens. RAG & 0.799 & 0.601 & 0.945 & 0.027 & 0.993 & 0.741 & 0.741 & 0.096 \\ 
            LLM-Ref & 0.964 & 0.637 & 0.954 & 0.195 & 0.999 & 0.654 & 0.561 & 0.422 \\
			\hline
		\end{tabular}
	}
	\caption{Metric Evaluation result comparison of LLM-Ref with Basic RAG, Parent Document Retriever RAG, and Ensemble Retrieval RAG for different LLMs. A higher value of a metric indicates better performance.}
	\label{tab:my-table_nlp_all}
\end{table*}


In the GPT-3.5 evaluation, LLM-Ref achieves the highest Answer Relevancy score of \(0.960\), markedly higher than Basic RAG (\(0.545\)), PDR RAG (\(0.619\)), and Ens. RAG (\(0.629\)). It also leads in Answer Correctness with \(0.555\), surpassing the others' range of \(0.412\) to \(0.471\). With an Answer Similarity of \(0.950\), LLM-Ref maintains a strong advantage over its peers, which hover between \(0.899\) and \(0.936\). These metrics confirm LLM-Ref’s superior capability to generate answers that are relevant and aligned with the provided context. Notably, while its Context Relevancy (\(0.157\)) is significantly higher than the others, it still lags behind in Context Recall, with scores slightly below those of Basic RAG (\(0.676\) vs \(0.665\)), but it compensates with a strong Faithfulness score of \(0.721\). The composite Ragas Score of \(0.389\) further highlights LLM-Ref’s overall effectiveness compared to the other methods, which range from \(0.052\) to \(0.143\).

For GPT-4o-mini, LLM-Ref retains its dominance with an Answer Relevancy score of \(0.953\), considerably higher than Basic RAG (\(0.765\)), PDR RAG (\(0.606\)), and Ens. RAG (\(0.857\)). Its Answer Correctness of \(0.575\) is on par with Ens. RAG (\(0.572\)) and significantly higher than other systems, reinforcing LLM-Ref's consistent accuracy. With the highest Answer Similarity (\(0.951\)) and a Ragas Score of \(0.413\), LLM-Ref continues to outperform other methods. However, its Context Recall (\(0.683\)) remains lower than PDR RAG (\(0.757\)) and Ens. RAG (\(0.689\)), suggesting room for improvement in extracting complete information from the context.

In the Llama 3.1-405b evaluation, LLM-Ref again exhibits superior performance with an Answer Relevancy score of \(0.958\) and an Answer Correctness score of \(0.556\), well above Basic RAG and PDR RAG, whose scores remain below \(0.650\). Its Answer Similarity of \(0.950\) and Faithfulness of \(0.564\) confirm that LLM-Ref provides high-quality, accurate responses while grounding its answers in relevant context. Although its Context Precision (\(0.987\)) is competitive, LLM-Ref still falls behind in Context Recall, with a score of \(0.650\) compared to Ens. RAG's \(0.725\). The Ragas Score for LLM-Ref is \(0.300\), much higher than Basic RAG (\(0.114\)) and PDR RAG (\(0.079\)).

Finally, with Claude 3.5 Sonnet, LLM-Ref maintains its strong performance across multiple metrics. It achieves the highest Answer Relevancy of \(0.964\), Answer Correctness of \(0.637\), and Answer Similarity of \(0.954\), outperforming other systems by substantial margins. While it continues to deliver accurate and relevant answers, its Context Recall score of \(0.654\) and Faithfulness score of \(0.561\) remain slightly lower compared to Ens. RAG (\(0.741\) for both). Despite this, LLM-Ref achieves the highest overall Ragas Score of \(0.422\), highlighting its superior performance in generating accurate and consistent answers across varied language models.

Across all LLM evaluations, LLM-Ref excels in delivering answers that are relevant, correct, and well-aligned with the input context. Its higher Ragas Scores across all models demonstrate its effectiveness in handling complex retrieval tasks, particularly in multi-document scenarios. However, the observed reductions in Context Recall and Faithfulness indicate potential areas where LLM-Ref could further improve, particularly in maximizing the utility of retrieved-context for single-document tasks.

\subsubsection{Stability Study}
As presented in Table~\ref{tab:my-table} and Table~\ref{tab:my-table_o1mini}, we provide comprehensive sets of evaluation metrics that underscore the effectiveness of our system. To assess our system's performance, it is essential to consider it holistically. Specifically, the context precision and context recall metrics are crucial for evaluating the retrieval stage, while faithfulness and answer relevancy are key indicators of the system's performance during the generation stage. Our metrics demonstrate superior performance across these stages.

In the content extraction stage, the process is deterministic; the system can either successfully extract text from a document or not. However, the summarization process introduces variability, as different summaries may be generated in each run, potentially impacting context extraction and the final response. To evaluate the stability of our system, we conducted multiple runs in a single-file scenario, with results indicating consistent performance with respect to Table~\ref{tab:my-table} given in the paper.

In the retrieval stage, unlike traditional RAG systems that typically select the top-k contexts, our approach involves retrieving all available contexts. This comprehensive retrieval method enhances the system's ability to generate accurate responses.

During the generation stage, we used a temperature setting of zero, ensuring that the model relies solely on the input context to generate responses, thereby minimizing randomness. We also experimented with varying the temperature parameter to observe its impact on response quality, as detailed in Table \ref{tab:temp-table}. We observed that as the temperature setting increases, the model tends to incorporate more of its pre-existing knowledge, which may include biases from its training data, potentially impacting the final Ragas score. The temperature parameter's influence on the model's output highlights the delicate balance between utilizing retrieved-context and minimizing reliance on potentially biased or extraneous information stored within the model. Consequently, adjusting the temperature parameter is crucial for maintaining the accuracy and integrity of the generated responses.

\begin{table*}[!ht]
\centering
\resizebox{\textwidth}{!}{
\begin{tabular}{cllllllll}
 \hline
        \multicolumn{8}{c}{Impact of temperature change in a single-source document setting}
        \\ 
\hline
  \multicolumn{1}{c}{\textbf{Temperature}} &
  \multicolumn{1}{c}{\textbf{\begin{tabular}[c]{@{}c@{}}Answer \\ Relevancy\end{tabular}}} &
  \multicolumn{1}{c}{\textbf{\begin{tabular}[c]{@{}c@{}}Answer \\ Correctness\end{tabular}}} &
  \multicolumn{1}{c}{\textbf{\begin{tabular}[c]{@{}c@{}}Context \\ Relevancy\end{tabular}}} &
  \multicolumn{1}{c}{\textbf{\begin{tabular}[c]{@{}c@{}}Context \\ Precision\end{tabular}}} &
  \multicolumn{1}{c}{\textbf{\begin{tabular}[c]{@{}c@{}}Context \\ Recall \end{tabular}}} &
  \multicolumn{1}{c}{\textbf{\begin{tabular}[c]{@{}c@{}}Faith \\ fulness\end{tabular}}} &
  \multicolumn{1}{c}{\textbf{\begin{tabular}[c]{@{}c@{}}Ragas \\ Score\end{tabular}}} \\ \hline
0.0  & 0.94 & 0.72 & 0.44 & 0.29 & 0.71 & 0.44 & 0.57 \\ 
0.05 & 0.94 & 0.71 & 0.40 & 0.27 & 0.74 & 0.40 & 0.54 \\ 
0.1  & 0.95 & 0.71 & 0.44 & 0.28 & 0.70 & 0.44 & 0.56 \\ 
0.15 & 0.93 & 0.67 & 0.35 & 0.24 & 0.65 & 0.37 & 0.49 \\ \hline

        \multicolumn{8}{c}{Performance variation across different runs in a single-source document setting for the same queries}
        \\ 
\hline
  \multicolumn{1}{c}{\textbf{Runs}} &
  \multicolumn{1}{c}{\textbf{\begin{tabular}[c]{@{}c@{}}Answer \\ Relevancy\end{tabular}}} &
  \multicolumn{1}{c}{\textbf{\begin{tabular}[c]{@{}c@{}}Answer \\ Correctness\end{tabular}}} &
  \multicolumn{1}{c}{\textbf{\begin{tabular}[c]{@{}c@{}}Context \\ Relevancy\end{tabular}}} &
  \multicolumn{1}{c}{\textbf{\begin{tabular}[c]{@{}c@{}}Context \\ Precision\end{tabular}}} &
  \multicolumn{1}{c}{\textbf{\begin{tabular}[c]{@{}c@{}}Context \\ Recall\end{tabular}}} &
  \multicolumn{1}{c}{\textbf{\begin{tabular}[c]{@{}c@{}}Faith \\ fulness\end{tabular}}} &
  \multicolumn{1}{c}{\textbf{\begin{tabular}[c]{@{}c@{}}Ragas \\ Score\end{tabular}}} \\ \hline
Run 1 & 0.95 & 0.70 & 0.35 & 0.24 & 0.71 & 0.41 & 0.52 \\ 
Run 2 & 0.94 & 0.72 & 0.44 & 0.29 & 0.71 & 0.44 & 0.57 \\ 
Run 3 & 0.94 & 0.70 & 0.38 & 0.25 & 0.68 & 0.45 & 0.54 \\ \hline
\end{tabular}
}
\caption{Stability study of our proposed approach.}
\label{tab:temp-table}
\end{table*}

\noindent These ablation studies highlight the robustness and adaptability of our system in generating precise and contextually relevant responses.


\subsection{Prompt Designs}
\label{promptdesigns}
In our tool, we employ a large language model (LLM) to determine contextual similarity. To find the relevant contexts, we utilize the following prompt (given in Figure~\ref{fig:prompt1}) which returns ‘True' when a paragraph is relevant to the query. This prompt instructs the LLM to evaluate a given paragraph in the context of a specific query, determining if it provides direct answers or significant contributions. Since we utilize entire paragraphs that convey specific concepts, the LLM can discern relevance to the query by understanding subtle nuances. By responding with ‘True' or ‘False', the model identifies relevant information without additional explanation, thereby enhancing the accuracy and efficiency of our tool.

\begin{figure}[!ht]
\centering
\begin{lstlisting}
You are an experienced researcher tasked with identifying relevant information.
Paragraph: {paragraph}
Query: {query}
Instructions: Determine whether the paragraph provides information that directly answers or significantly contributes to the query. 
If the paragraph is relevant to the query, respond with 'True'. If it is not relevant, respond with 'False'. Provide no additional explanation. 
\end{lstlisting}
\caption{Prompt to find relevant contexts to a query.}
\label{fig:prompt1}
\end{figure}

To address challenges associated with long contexts, we employ an iterative approach to output generation. Initially, a response is generated using the first context and query, utilizing the LLM prompt provided in Figure~\ref{fig:prompt2}. 

\begin{figure}[!ht]
\centering
\begin{lstlisting}
You are a researcher writing a research paper. 
**Paragraph**: {paragraph}                        
**Query**: {query}
**Instructions**: Summarize and synthesize the provided paragraph to create a cohesive and informative paragraph that addresses the query. 
Ensure the synthesis uses the vocabulary and writing style of the original paragraph to maintain a natural and consistent tone.
\end{lstlisting}
\caption{Prompt used to generate the response based on the context for query.}
\label{fig:prompt2}
\end{figure}

This prompt (given in Figure~\ref{fig:prompt2}) directs the LLM to summarize and synthesize the paragraph to address the query coherently. By preserving the original vocabulary and style, the LLM ensures a natural and consistent tone. This iterative approach manages long contexts and enhances the relevance and cohesiveness of the responses, improving our tool's efficiency and accuracy. After the initial response is generated, subsequent responses are refined by incorporating later contexts using the following prompt (shown in Figure~\ref{fig:prompt3}). This iterative approach not only enhances the comprehensiveness of the synthesized output but also helps in mitigating any errors present in the earlier responses.

\begin{figure}[!ht]
\centering
\begin{lstlisting}
You are a researcher writing a research paper. 
**Existing Synthesis**: {response}                            
**New Paragraph**: {paragraph}
**Query**: {query}
**Instructions**: Integrate the information from the new paragraph into the existing synthesis to create a cohesive and informative paragraph that addresses the query. 
Ensure the synthesis uses the vocabulary and writing style of the original paragraphs to maintain a natural and consistent tone.
\end{lstlisting}
\caption{Prompt used to integrate new context into existing responses.}
\label{fig:prompt3}
\end{figure}
This prompt (given in Figure~\ref{fig:prompt3}) guides the LLM to integrate new paragraph information into the existing synthesis, maintaining coherence, relevance, and a consistent tone, while iteratively refining responses to address long context complexities and improve the tool's accuracy and cohesiveness.

Figure~\ref{fig:prompt4} shows a prompt directing the LLM to match each line of a synthesized result with the most relevant source lines from the provided paragraphs. The output lists only the precisely relevant source lines, enhancing the traceability and transparency of the synthesis process by clarifying the origins of each part of the synthesized result.

\begin{figure}[!ht]
\centering
\begin{lstlisting}
For a given synthesized result based on some source paragraphs, find the relevant source lines that are most relevant to each line of the synthesized result. 
Synthesized result: {synthesized_result}.  
Source Paragraphs:  {context}. 
Just provide the source lines for each line of synthesized result, for example: Synthesized Line: ... Corresponding Source Line: ...   Do not add explanation and source lines if they are not exactly relevant.
\end{lstlisting}
\caption{Prompt for identifying the most relevant source lines for each line in a synthesized result.}
\label{fig:prompt4}
\end{figure}

Figure~\ref{fig:prompt5} presents a prompt to generate questions by synthesizing information from at least two of three provided documents. The prompt requires formulating questions, including exact original context texts, and providing answers, all in a specified Python format. This ensures the integrity of the original contexts for evaluation. Questions are generated until a certain number of unique questions are produced, enhancing the tool's ability to synthesize information accurately across multiple documents.

\begin{figure}[!ht]
\centering
\begin{lstlisting}
You are an expert research scientist.
Instructions: Create a list of 150 questions (max 5 at a time) that require using information from all three provided input documents (or at least two of the input documents). For each question, please include the following details:

Question: Formulate a question that integrates information from multiple documents.
Original Context Texts: Provide the exact contexts from the documents that were used to create the question, without any alterations.
Answer: Provide an answer for a research article derived from the original context texts.
Ensure that each question requires the synthesis of information from multiple documents. Maintain the integrity of the original context texts as they will be used later for evaluation purposes.
Return the response in the following python format: 
data = [
    {
        "question": "Question 1",
        "context": ["Context 11", "Context 12"],
        "ground_truth": "Answer 1"
    },
    {
        "question": "Question 2",
        "context": ["Context 21", "Context 22"],
        "ground_truth": "Answer 2"
    },]

    
Please keep generating only if it is possible to generate unique questions that you did not generate them before. Generate 5 questions at a time. I want a total 150 questions.

\end{lstlisting}
\caption{Prompts for generating Question-Context-Answer pair from source documents.}
\label{fig:prompt5}
\end{figure}

\subsection{Ragas Evaluation Metrics}
\label{ragas_appendix}
The Ragas score is computed by calculating the harmonic mean of Faithfulness (FF), Answer Relevancy (AR), Context Precision (CP), and Context Recall (CR).

\begin{equation}
\text{Ragas Score} = \frac{4}{\frac{1}{\text{FF}} + \frac{1}{\text{AR}} + \frac{1}{\text{CP}} + \frac{1}{\text{CR}}}
\end{equation}

\noindent In this equation, FF stands for Faithfulness, AR represents Answer Relevancy, CP is Context Precision, and CR denotes Context Recall. In the RAGs framework, Faithfulness and Answer Relevancy assess the accuracy of content generation, while Context Precision and Context Recall evaluate the effectiveness of information retrieval. Therefore, the Ragas score ensures a robust assessment of both generation and retrieval processes in RAGs.

\textbf{Faithfulness (FF):} The Faithfulness score measures how relevant the statements in an answer are to the provided context. Scores for this metric range from 0 to 1, with higher scores indicating better alignment and performance. The calculation process, as defined by the Ragas framework, involves three key steps: first, extracting statements from the generated answers; second, determining the contextual relevance of these statements using the LLM; and third, calculating the Faithfulness score by dividing the number of context-relevant statements by the total number of statements. This score provides a quantifiable measure of how faithfully the model's answers reflect the original context. It is calculated as:

\begin{equation}
\text{FF} = \frac{\text{NCS}}{\text{TS}}
\end{equation}

\noindent Here, NCS refers to the Number of Context-Relevant Statements, and TS represents the Total Statements in the Answer.

\noindent \textbf{Answer Relevancy (AR):} The Answer Relevance metric evaluates how closely the answers generated by a Language Learning Model (LLM) align with the original questions posed. Answers that are incomplete or redundant receive lower scores, with scores ranging from 0 to 1, where higher scores indicate better performance. The Ragas framework calculates this metric through a three-step process: first, generating pseudo-questions from both the context and the generated answer; second, calculating the cosine similarity between the original question and each pseudo-question; and third, computing the average of these cosine similarities. This average provides a quantitative measure of how relevant the generated answers are to the original questions.

\begin{equation}
\text{AR} = \frac{\sum \text{CS}}{\text{NPQ}}
\end{equation}

\noindent In this context, CS denotes Cosine Similarities between pseudo-questions and the original question, and NPQ stands for the Number of Pseudo-Questions.

\vspace{0.5cm}

\noindent \textbf{Context Precision (CP):} The Context Precision metric measures how effectively a Language Learning Model (LLM) retrieves the necessary contextual information required to accurately answer a question. Scores for this metric range from 0 to 1, with higher scores indicating better retrieval performance. According to the Ragas framework, Context Precision is calculated through a two-step process: first, determining the relationship between each retrieved-context and the original question using the LLM, where the context is marked as either relevant (Yes) or not (No); and second, computing the Mean Average Precision (mAP) across all retrieved contexts. This score indicates how accurately the model retrieves relevant information to support its answers.

\begin{equation}
\text{CP} = \text{mAP}
\end{equation}

\noindent \textbf{Context Recall (CR):} The Context Recall metric evaluates how well the context retrieved by a Language Learning Model (LLM) matches the Ground Truth, indicating the completeness of the information retrieval. Scores range from 0 to 1, with higher scores reflecting better performance. The Ragas framework computes this metric through a three-step process: first, splitting the Ground Truth into individual sentences; second, determining the relationship between each sub-Ground Truth sentence and the retrieved context using the LLM, marking each as either relevant (Yes) or not (No); and third, calculating the Context Recall score by dividing the number of context-relevant Ground Truth sentences by the total number of Ground Truth sentences. This score helps in quantifying how thoroughly the model's retrieved-context covers the Ground Truth.

\begin{equation}
\text{CR} = \frac{\text{NGTS}}{\text{TGS}}
\end{equation}

\noindent Here, NGTS stands for the Number of Ground Truth Sentences inferred from the given contexts, and TGS represents the Total Ground Truth Sentences.

\subsection{Examples of Query-Answer Pairs}
We present additional query-answer pairs with fine-grained references extracted from LLM-Ref when different LLMs are utilized. This demonstrates the compatibility of our tool across different LLMs. 
\begin{figure*}[!ht]
	\centering
	\begin{tabular}{p{15.7cm}}
		\hline
		\footnotesize
		Query: How does data augmentation improve speech-to-speech translation? \\ \hline
		
		\footnotesize Answer: Data augmentation plays a crucial role in enhancing speech-to-speech translation (S2ST) systems by addressing the challenges posed by data scarcity [1-8]. In the context of end-to-end S2T models, effective training strategies such as multitask learning and model pre-training are complemented by data augmentation techniques, which help to enrich the training dataset[1-4, 9-13]. This enrichment allows the models to better generalize and improve their performance, particularly in mitigating the error propagation issues commonly encountered in traditional ASR+MT systems[1-3, 14-16]. By diversifying the training data, data augmentation facilitates the synthesis of para-linguistic information, such as prosody and word-level emphasis, which are essential for producing more natural and contextually appropriate translations[1-3, 17-19]. Recent empirical findings further demonstrate that augmenting supervised S2ST data with directly mined S2ST data, particularly when combined with speech normalization techniques, can significantly enhance the performance of textless S2ST systems[2-3]. For instance, the integration of mined data has shown to yield substantial improvements in BLEU scores across various language pairs, indicating that the additional data not only enriches the training set but also aids in capturing variations in speech that are difficult to model without such information.Moreover, the application of data augmentation techniques, such as utilizing ASR, MT, and TTS models to create weakly supervised data, has been shown to effectively tackle the data scarcity issue in S2ST. By leveraging the S2UT framework, researchers have demonstrated that large-scale self-supervised pre-training with monolingual speech and text data, combined with data augmentation, can significantly enhance S2ST model training [3,12]. Specifically, the use of a wav2vec 2.0 speech encoder and an mBART text decoder trained with discrete units extracted from unlabeled speech data has proven effective [2-3, 14-16].	The incorporation of weakly supervised data from ASR speech further increases the size of the parallel S2ST training data, leading to improved model performance[2,3]. For example, experiments have shown that with a pre-trained wav2vec 2.0 encoder and a randomly initialized decoder, substantial BLEU gains can be achieved, indicating that data augmentation not only enriches the training dataset but also enhances the model's ability to generalize across different setups, including low-resource scenarios[2-3]. Thus, through these enhancements, data augmentation significantly contributes to the robustness and efficacy of S2ST systems, enabling them to leverage both supervised and mined data for improved translation quality[1-2].\\ \hline
		
		\footnotesize
		References:
		\begin{enumerate}[nosep, label={[{\arabic*}]}]
			\item A. Lee et al., “Direct speech-to-speech translation with discrete units,” arXiv, 2021.
			\item A. Lee et al., “Textless speech-to-speech translation on real data,” arXiv, 2021.
			\item S. Popuri et al., “Enhanced Direct Speech-to-Speech Translation Using Self-supervised Pre-training and Data Augmentation,” arXiv, 2022.
			\item Ye Jia et al., Direct speech-to-speech translation with a sequence-to-sequence model. Proc. Interspeech 2019.
			\item R. J. Weiss, J. Chorowski et al., “Sequence-to-sequence models
			can directly translate foreign speech,” Proc. Interspeech, 2017.
			
			\item J. Pino, Q. Xu et al., “Self-training for end-to-end speech translation,”
			Proc. Interspeech, 2020.
			
			\item J. Kahn, A. Lee et al., “Self-training for end-to-end speech recognition,”
			in ICASSP, 2020.
			
			\item  T. Hayashi, S. Watanabe et al., “Back-translation-style data augmentation
			for end-to-end asr,” in SLT, 2018.
			
			\item Alexandre Bérard, Olivier Pietquin, Christophe Servan,
			and Laurent Besacier. 2016. Listen and translate: A
			proof of concept for end-to-end speech-to-text translation.
			arXiv preprint arXiv:1612.01744.
			
			\item Parnia Bahar, Tobias Bieschke, and Hermann Ney.
			2019. A comparative study on end-to-end speech
			to text translation. In 2019 IEEE Automatic Speech
			Recognition and Understanding Workshop (ASRU),
			pages 792–799. IEEE.
			
			\item Xian Li, Changhan Wang, Yun Tang, Chau Tran,
			Yuqing Tang, Juan Pino, Alexei Baevski, Alexis
			Conneau, and Michael Auli. 2021. Multilingual
			speech translation from efficient finetuning of pretrained
			models. In Proceedings of the 59th Annual
			Meeting of the Association for Computational Linguistics
			and the 11th International Joint Conference on Natural Language Processing (Volume 1: Long
			Papers), pages 827–838.

			\item 	C. Zhang, X. Tan et al., “UWSpeech: Speech to speech translation
			for unwritten languages,” arXiv:2006.07926, 2020.
			
			\item A. Polyak, Y. Adi et al., “Speech resynthesis from discrete
			disentangled self-supervised representations,” arXiv:2104.00355,
			2021.
			
			\item M. C. Stoian, S. Bansal et al., “Analyzing asr pretraining for lowresource
			speech-to-text translation,” in ICASSP, 2020.
			
			\item A. Baevski, Y. Zhou et al., “wav2vec 2.0: A framework for selfsupervised
			learning of speech representations,” Neurips, vol. 33,
			pp. 12 449–12 460, 2020.
			\item  Y. Liu, J. Gu et al., “Multilingual denoising pre-training for neural
			machine translation,” Transactions of the Association for Computational
			Linguistics, vol. 8, pp. 726–742, 2020.
			
			\item Q. T. Do, et al., “Toward expressive speech translation: A unified sequence-to-sequence LSTMs approach for translating words and emphasis.” In INTERSPEECH, 2017.
			\item P. D. Aguero, et al., “Prosody generation for speech-to-speech translation.” In 2006 IEEE International Conference on Acoustics Speech and Signal Processing Proceedings, volume 1, pages I–I, 2006.
			\item G. K. Anumanchipalli et al., “Intent transfer in speech-to-speech machine translation.” In 2012 IEEE Spoken Language Technology Workshop (SLT), 2012.
		
		\end{enumerate} \\
		\hline
	\end{tabular}
	\caption{Fine-grained reference samples generated by LLM-Ref when GPT-4o-mini is used as the LLM.}
	\label{tab:fine_4omini}
\end{figure*}

\begin{figure*}[!ht]
	\centering
	\begin{tabular}{p{15.7cm}}
		\hline
		\footnotesize
		Query: Literature review on Direct speech-to-speech translation? \\ \hline
		
		\footnotesize Answer: Direct speech-to-speech translation (S2ST) aims at translating speech from one language into speech in another language without relying on text generation as an intermediate step. Compared to conventional cascaded approaches, which take advantage of automatic speech recognition (ASR), machine translation (MT) or end-to-end speech-to-text translation (S2T) followed by text-to-speech synthesis (TTS), direct S2ST has the advantage of faster inference and can support translation between languages without text writing systems [1-12]. Recent studies have proposed various approaches to direct S2ST, including self-supervised speech encoder pre-training with unlabeled speech data, and data augmentation techniques that benefit S2T model training [1-3, 13-20,.]. For instance, [4] proposes to apply a self-supervised speech encoder pre-trained on unlabeled speech to convert target speech into discrete units and build a speech-to-unit translation (S2UT) model for direct S2ST [1-3, 5,  9-11,  15, 17, 20,.]. Self-supervised discrete targets can disentangle linguistic content from speaker identity and prosodic information in speech, and enable opportunities for applying techniques from speech-to-text model training, such as ASR and S2T, to direct S2ST[1-5, 10, 13 20-23]. Moreover, [6] shows that incorporating weakly supervised training data from ASR speech can bring significant improvements to direct S2ST models[1-5, 13, 15, 20-22]. Our work builds upon these advances and explores the effectiveness of self-supervised pre-training and data augmentation for direct S2ST models.\\ \hline
		\footnotesize
		References:
		\begin{enumerate}[nosep, label={[{\arabic*}]}]
			\item A. Lee et al., “Direct speech-to-speech translation with discrete units,” arXiv, 2021.
			\item A. Lee et al., “Textless speech-to-speech translation on real data,” arXiv, 2021.
			\item S. Popuri et al., “Enhanced Direct Speech-to-Speech Translation Using Self-supervised Pre-training and Data Augmentation,” arXiv, 2022.
			
			\item A. Tjandra et al., “Speech-to-speech translation between untranscribed unknown languages.” In 2019 IEEE Automatic Speech Recognition and Understanding Workshop (ASRU), 2019.
			
			\item C. Zhang, X. Tan et al., “UWSpeech: Speech to speech translation for unwritten languages,” arXiv:2006.07926, 2020.
			
			\item Lavie et al., “JANUS-III: Speech-to-speech translation in multiple languages.” In 1997 IEEE International Conference on Acoustics, Speech, and Signal Processing.
			
			\item S. Nakamura, The ATR multilingual speech-to-speech translation system. IEEE Transactions on Audio, Speech, and Language Processing, 2006.
			
			\item Alexandre Bérard et. al.. 2016. Listen and translate: A proof of concept for end-to-end speech-to-text translation. arXiv preprint arXiv:1612.01744.
			
			\item Ye Jia et al., Direct speech-to-speech translation with a sequence-to-sequence model. Proc. Interspeech 2019.
			
			\item Ye Jia et al., Translatotron 2: Robust direct speech-to-speech translation. arXiv 2021.
			
			\item Takatomo Kano et. al. 2021. Transformer-based direct speech-to-speech
			translation with transcoder. In 2021 IEEE Spoken Language Technology Workshop (SLT), pages 958–965. IEEE.
			
			\item Aaron van den Oord et. al., 2017. Neural discrete representation
			learning. In Proceedings of the 31st International Conference on Neural Information Processing Systems,	pages 6309–6318.
			
			\item Shu-wen Yang et al. 2021. SUPERB: Speech processing
			universal performance benchmark. arXiv preprint
			arXiv:2105.01051.

			\item A. Baevski, Y. Zhou et al., “wav2vec 2.0: A framework for selfsupervised
			learning of speech representations,” Neurips, vol. 33,
			pp. 12 449–12 460, 2020.

			\item W. Hsu, HuBERT: Self-supervised speech representation learning by masked prediction of hidden units. arXiv preprint arXiv:2106.07447.

			\item Zhiyun Fan et. al. 2020. Exploring wav2vec 2.0 on speaker verification
			and language identification. arXiv preprint arXiv:2012.06185.
			
			\item R. J. Weiss, J. Chorowski et al., “Sequence-to-sequence models
			can directly translate foreign speech,” Proc. Interspeech, 2017. 
			
			\item Parnia Bahar et al.
			2019. A comparative study on end-to-end speech
			to text translation. In 2019 IEEE Automatic Speech
			Recognition and Understanding Workshop (ASRU),
			pages 792–799. IEEE.
			
			\item Xian Li et. al. 2021. Multilingual
			speech translation from efficient finetuning of pretrained
			models. In Proceedings of the 59th Annual
			Meeting of the Association for Computational Linguistics
			and the 11th International Joint Conference on Natural Language Processing (Volume 1: Long
			Papers), pages 827–838.
			
			\item Kushal Lakhotia et al. 2021. Generative spoken language
			modeling from raw audio. arXiv preprint
			arXiv:2102.01192.

			\item A. Polyak, Y. Adi et al., “Speech resynthesis from discrete
			disentangled self-supervised representations,” arXiv:2104.00355,
			2021.
			
			\item E. Kharitonov, A. Lee et al., “Text-free prosody-aware generative
			spoken language modeling,” arXiv:2109.03264, 2021.
			
			\item F. Kreuk et al., “Textless speech emotion
			conversion using decomposed and discrete representations,”
			arXiv:2111.07402, 2021.
\end{enumerate} \\
		\hline
	\end{tabular}
	\caption{Fine-grained reference samples generated by LLM-Ref when Llama is used as the LLM.}
	\label{tab:fine_llama}
\end{figure*}
\end{document}